\documentclass{article}

\usepackage{microtype}
\usepackage{graphicx}
\usepackage{subfigure}
\usepackage{booktabs} 

\usepackage{hyperref}

\usepackage[accepted]{icml2024}

\usepackage{amsmath}
\usepackage{amssymb}
\usepackage{mathtools}
\usepackage{amsthm}

\usepackage{amsfonts}
\usepackage{bm}
\usepackage{multirow}
\usepackage{multicol}
\usepackage{caption}
\usepackage{balance}
\usepackage{xcolor}
\usepackage{rotating}
\usepackage{diagbox}
\usepackage{bbding}
\usepackage{svg}

\def\method{HEAL}

\usepackage[capitalize,noabbrev]{cleveref}

\theoremstyle{plain}

\theoremstyle{definition}

\theoremstyle{remark}

\usepackage[textsize=tiny]{todonotes}

\icmltitlerunning{Hypergraph-enhanced Dual Semi-supervised Graph Classification}

\begin{document}

\twocolumn[
\icmltitle{Hypergraph-enhanced Dual Semi-supervised Graph Classification}




\begin{icmlauthorlist}
\icmlauthor{Wei Ju}{1}
\icmlauthor{Zhengyang Mao}{1}
\icmlauthor{Siyu Yi$^*$}{2}
\icmlauthor{Yifang Qin}{1}
\icmlauthor{Yiyang Gu}{1}
\icmlauthor{Zhiping Xiao}{3}
\icmlauthor{Yifan Wang}{4}\\
\icmlauthor{Xiao Luo}{3}
\icmlauthor{Ming Zhang$^*$}{1}
\end{icmlauthorlist}

\icmlaffiliation{1}{School of Computer Science, National Key Laboratory for Multimedia Information Processing, Peking University-Anker Embodied AI Lab, Peking University, China}
\icmlaffiliation{2}{School of Statistics and Data Science, Nankai University, China}
\icmlaffiliation{3}{Department of Computer Science, University of California, Los Angeles, USA}
\icmlaffiliation{4}{School of Information Technology $\&$ Management, University of International Business and Economics, China}

\icmlcorrespondingauthor{Siyu Yi}{siyuyi@mail.nankai.edu.cn}
\icmlcorrespondingauthor{Ming Zhang}{mzhang\_cs@pku.edu.cn}

\icmlkeywords{Machine Learning, ICML}

\vskip 0.3in
]



\printAffiliationsAndNotice{}  

\begin{abstract}
In this paper, we study semi-supervised graph classification, which aims at accurately predicting the categories of graphs in scenarios with limited labeled graphs and abundant unlabeled graphs. Despite the promising capability of graph neural networks (GNNs), they typically require a large number of costly labeled graphs, while a wealth of unlabeled graphs fail to be effectively utilized. Moreover, GNNs are inherently limited to encoding local neighborhood information using message-passing mechanisms, thus lacking the ability to model higher-order dependencies among nodes. To tackle these challenges, we propose a \textbf{H}ypergraph-\textbf{E}nhanced Du\textbf{AL} framework named HEAL for semi-supervised graph classification, which captures graph semantics from the perspective of the hypergraph and the line graph, respectively. Specifically, to better explore the higher-order relationships among nodes, we design a hypergraph structure learning to adaptively learn complex node dependencies beyond pairwise relations. Meanwhile, based on the learned hypergraph, we introduce a line graph to capture the interaction between hyperedges, thereby better mining the underlying semantic structures. Finally, we develop a relational consistency learning to facilitate knowledge transfer between the two branches and provide better mutual guidance. Extensive experiments on real-world graph datasets verify the effectiveness of the proposed method against existing state-of-the-art methods.
\end{abstract}


\section{Introduction}

Graph classification, which involves identifying the class labels of graphs, is a significant problem with diverse practical applications in various fields. Data originating from domains such as bioinformatics, chemoinformatics, and social network analysis, can naturally be represented as graphs. For example, molecules in chemoinformatics can be represented as graphs by viewing atoms as nodes and chemical bonds between pairs of atoms as edges. The objective of this task is to effectively recognize the class label of each graph, such as predicting the quantum mechanical properties~\cite{hao2020asgn,ju2023few} and assessing the functionality of chemical compounds~\cite{kojima2020kgcn}.

To solve this problem, early methods leverage the idea of graph kernels that compute a similarity measure between graphs by comparing their substructures~\cite{kashima2003marginalized,shervashidze2009efficient,shervashidze2011weisfeiler}, and have been proven to effectively capture the structural properties of graphs. Despite their efficacy, graph kernels may face challenges in scalability and computational efficiency when dealing with large datasets or complex graphs. Recently, graph neural networks (GNNs)~\cite{kipf2017semi,ju2024comprehensive} have emerged as a prominent and powerful paradigm for graph classification~\cite{mao2023rahnet,yi2023towards,luo2023towards}. The key idea of GNNs is to learn effective graph representations by iteratively aggregating information from neighboring nodes~\cite{gilmer2017neural}, which have achieved remarkable success.

However, most prevailing methods typically follow the framework of supervised learning, which demands a substantial amount of labeled graphs to train GNN models. However, in the field of graph analytics, obtaining labeled graphs can be a costly endeavor. Annotating graph data requires expert domain knowledge, manual efforts, and often extensive human involvement, making the process time-consuming and expensive. For instance, molecular labels are often acquired through costly Density Functional Theory (DFT) calculations or generated from complex experiments in the field of chemistry~\cite{hao2020asgn,ju2023few}. This scarcity of labeled graphs and the high cost of annotation pose significant challenges for developing accurate and robust GNNs for graph classification.

This inspires us to study semi-supervised graph classification, where we leverage both labeled and unlabeled graphs. Despite the unavailability of properties (\emph{i.e.}, labels) in the unlabeled graphs, their structures contain valuable information that could potentially enrich the capabilities of GNNs if utilized effectively. Actually, there are several approaches along this line~\cite{li2019semi,hao2020asgn,luo2022dualgraph,ju2022kgnn,ju2023tgnn}. ASGN~\cite{hao2020asgn} adopts a teacher-student framework to fully utilize labeled and unlabeled graphs. DualGraph~\cite{luo2022dualgraph} incorporates contrastive learning to encourage the consistency of unlabeled graphs in a dual manner. KGNN~\cite{ju2022kgnn} and TGNN~\cite{ju2023tgnn} unify the GNNs and graph kernels in a semi-supervised framework.

Despite the encouraging performance achieved by existing methods, they still suffer from two key limitations. \emph{First}, GNNs are restricted to capturing only low-order local neighborhood information and struggle to model high-order dependencies between nodes. For instance, in chemoinformatics, GNNs may be unable to effectively capture the complex interactions and long-range dependencies between atoms in a molecule, thereby potentially limiting their ability to accurately predict properties like molecular activity or toxicity. \emph{Second}, the utilization of unlabeled graphs remains underexploited, despite their containing valuable structural information. The unlabeled graphs can act as a regularizer, facilitating the exploration of intrinsic graph semantics, even in the presence of a scarce amount of labeled graphs. For example, in social network analysis, there might be a large amount of unlabeled user interaction graphs, yet these graphs hold insightful community structures and social relationships. As such, we are highly desired to look for an approach that is able to better capture high-order dependencies among nodes and meanwhile sufficiently leverage the unlabeled graphs to overcome the scarcity of labeled graphs.

To address these challenges, in this paper we propose a \textbf{H}ypergraph-\textbf{E}nhanced Du\textbf{AL} framework named HEAL for semi-supervised graph classification. The key idea of \method{} is to capture graph semantics from the perspective of the hypergraph and the line graph, respectively. Specifically, to explore the intricate interdependencies among nodes, we develop a learnable hypergraph structure learning, which possesses the remarkable ability to adaptively acquire higher-order node relationships beyond pairwise connections, and is more flexible to model complex data structures than pre-defined hypergraph construction. Moreover, due to the presence of higher-order semantic interactions in complex data structures, we hence leverage the learned hypergraph to introduce a line graph, effectively capturing the interactions between hyperedges, thus unlocking deeper insights into the underlying semantic structures of graphs. Finally, since the hypergraph and the line graph explore graph semantics at different levels of higher-order structures, it is crucial to jointly train these two branches to enable mutual knowledge transfer between them. We thus present relational consistency learning, in which two branches are required to produce consistent similarity scores for each unlabeled graph. By encouraging the consistency between two similarity distributions, our method effectively enhances the potential of the model by fully using unlabeled graphs, thereby better serving the semi-supervised graph classification. Experiments validate the effectiveness of our proposed model \method{}.

    
\section{Problem Definition \& Preliminaries}

\begin{figure*}[t!]
    \centering
    \includegraphics[width=0.9\textwidth]{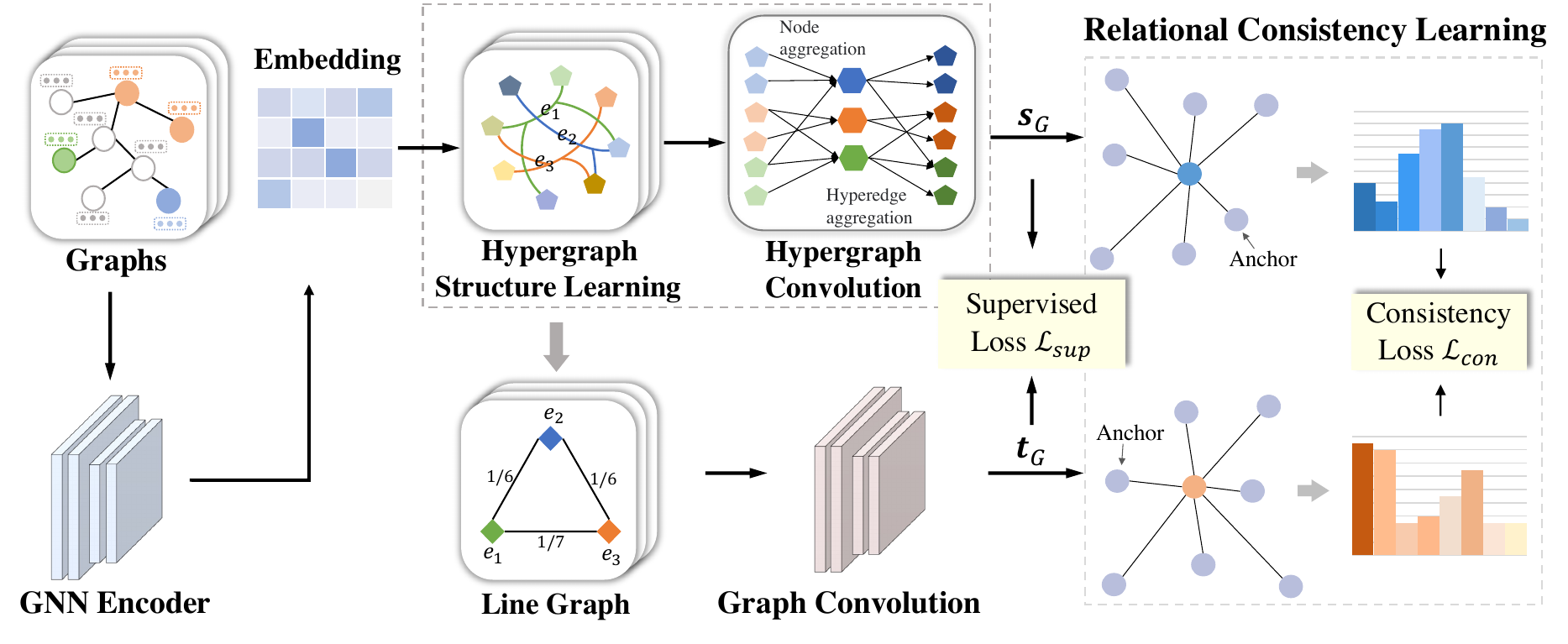}
    \caption{{Illustration of the proposed framework \method{}. 
    }}
    \label{fig:framework}
\end{figure*}


\noindent\textbf{Definition 1.}
\emph{A \textbf{graph} can be defined as $G = (V, E, \mathbf{X}, \mathbf{y})$, where $V$ is a set of nodes, and $E$ is a set of edges. $\mathbf{X}\in R^{|V|\times d_f}$  denotes the feature matrix of nodes, where $d_f$ is the dimension of features. $\mathbf{y}$ is the class label of graph $G$ and $\mathbf{A}\in \{0, 1\}^{|V|\times |V|}$ represents the adjacency matrix.}

\smallskip
\noindent\textbf{Definition 2.}
\emph{A \textbf{hypergraph} is a generalization of a graph where edges are allowed to connect more than two nodes. Formally, a hypergraph is represented as $H = (V, E_H)$, where $V$ is the node set same as in graph $G$, and $E_H$ is the set of hyperedges, which can contain any number of nodes. Each hyperedge $e_h \in E_H$ is a subset of the node set $V$.}

\smallskip
\noindent\textbf{Definition 3.}
\emph{A \textbf{line graph} of the hypergraph is defined as a graph $L(H) = (V_L, E_L)$, where each node $v_l \in V_L$ corresponds to an edge in $H$, and two nodes in $V_L$ are adjacent in $L(H)$ if and only if the corresponding edges in $H$ share a common node~\cite{whitney1992congruent}. Formally, $V_L=\left\{v_l: v_l \in E_H \right\}$, and $E_L=\left\{\left(v_{l_i}, v_{l_j}\right): v_{l_i}\right.$, $\left.v_{l_j} \in E_H, \left|v_{l_i} \cap v_{l_j}\right| \geq 1\right\}$. The weight of each edge $W_{i, j}$ is assigned to $W_{i, j}=\left|v_{l_i} \cap v_{l_j}\right| /\left|v_{l_i} \cup v_{l_j}\right|$.} 



\smallskip
\noindent\textbf{Semi-supervised Graph Classification.}
Given a set of graphs $\mathcal{G} = \{\mathcal{G}^L, \mathcal{G}^U\}$, in which $\mathcal{G}^L = \left\{G_{1}, \cdots, G_{\left|\mathcal{G}^{L}\right|}\right\}$ are labeled graphs and $\mathcal{G}^{U}=$ $\left\{G_{\left|\mathcal{G}^{L}\right|+1}, \cdots, G_{\left|\mathcal{G}^{L}\right|+\left|\mathcal{G}^{U}\right|}\right\}$ are unlabeled graphs. the problem of semi-supervised graph classification can be defined as learning a mapping function from graphs to class labels $f: \mathcal{G} \rightarrow \mathcal{Y}$ to predict the labels of $\mathcal{G}^{U}$, where $\mathcal{Y}$ represents the labels corresponding to $\mathcal{G}$.

\smallskip
\noindent\textbf{GNN-based Encoder.} The general mechanism of GNNs is to iteratively update node
embeddings by aggregating the information of its neighbor nodes via message-passing \cite{gilmer2017neural}. Formally, the node embeddings $\mathbf{H}  = [\mathbf{h}_{1}, \mathbf{h}_{2}, \dots, \mathbf{h}_{|V|}]^{\top} \in \mathbb{R}^{|V| \times d}$ can be updated as:
\begin{equation}
\label{eq:gnn}
\mathbf{H}= \sigma(\hat{\mathbf{A}} \mathbf{X} \mathbf{W}), \quad \hat{\mathbf{A}}=\tilde{\mathbf{D}}^{-\frac{1}{2}} \tilde{\mathbf{A}} \tilde{\mathbf{D}}^{-\frac{1}{2}},
\end{equation}
where $\tilde{\mathbf{A}}=\mathbf{A}+\mathbf{I}$, $\tilde{\mathbf{D}}$ is the degree matrix of $\tilde{\mathbf{A}}$, $\mathbf{W}$ is the trainable weight matrix, and $\sigma(\cdot)$ is the activation function. Then the whole graph representation $\mathbf{h}_G$ can be computed based on all node embeddings as:
\begin{equation}
    \mathbf{h}_G=\sum\nolimits_{i=1}^{|V|} \mathbf{h}_{i}.
\end{equation}


\section{Methodology}
\label{sec::model}


In this section, we introduce our \method{} framework for semi-supervised graph classification, which captures graph semantics from both the hypergraph and line graph perspectives. \method{} consists of three modules: hypergraph high-order dependency learning, graph convolution on the line graph, and relational consistency learning. Figure~\ref{fig:framework} provides an overview of the whole framework. 

\subsection{Hypergraph High-order Dependency Learning}

GNNs have achieved significant success in learning expressive representations. However, they are inherently limited to capturing only local neighborhood information via message-passing mechanisms~\cite{gilmer2017neural} and cannot effectively capture higher-order substructures. This limitation is crucial since many real-world graph data exhibit complex hierarchical relationships that extend beyond immediate neighbors. To address this issue, we propose using hypergraphs to overcome the aforementioned limitation of GNNs. Hypergraphs~\cite{feng2019hypergraph} provide a more powerful framework for modeling higher-order dependencies and interactions among nodes, enabling us to better capture the rich structural information present in the graphs.

\noindent\textbf{Hypergraph Structure Learning.}
Existing approaches typically construct hypergraphs using predefined criteria based on distances~\cite{yu2012adaptive}, representations~\cite{wang2015visual}, or attributes~\cite{huang2015learning}. However, these methods may suffer from sub-optimal performance and high computational costs due to their lack of flexibility. To overcome these limitations, we develop a flexible way to parameterize a learnable hypergraph structure, being optimized jointly with the network parameters. Nevertheless, directly learning a dense adjacency matrix could incur excessive computational overhead, so we instead adopt a low-rank strategy to efficiently model the hypergraph structural matrix $\mathbf{\Lambda} \in \mathbb{R}^{|V|\times k}$, where $k$ represents the number of hyperedges, calculated as:
\begin{equation}
    \mathbf{\Lambda}=\mathbf{H} \cdot \mathbf{W},
\end{equation}
where $\mathbf{H} \in \mathbb{R}^{|V|\times d}$ is the hidden embedding matrix derived from the GNN-based encoder, and $d$ is the dimension of hidden embeddings. We introduce a learnable weight matrix $\mathbf{W} \in \mathbb{R}^{d \times k}$ to model the hyperedges. As a result, learning the hypergraph structural matrix requires only $\mathcal{O}(k \times d)$ time complexity $(d<<|V|)$, which significantly improves model efficiency without compromising performance. 

\noindent\textbf{Hypergraph Convolution.}
After obtaining a flexible hypergraph, we can effectively capture higher-order dependencies among nodes. To achieve this, we design a hypergraph convolution to learn high-level node representations. First, we learn hyperedge embeddings by aggregating neighbor nodes connected in the hypergraph. Then, the acquired hyperedge embeddings are leveraged to perform higher-order updates on the node representations. Technically, the hyperedge embedding matrix $\mathbf{R} \in \mathbb{R}^{k \times d}$ is computed as follows:
\begin{equation}
    \mathbf{R}=\sigma\left(\mathbf{U} \mathbf{\Lambda}^{\top} \mathbf{H}\right)+\mathbf{\Lambda}^{\top} \mathbf{H},
\end{equation}
where $\mathbf{U} \in \mathbb{R}^{k \times k}$ denotes the additional trainable matrix, and $\sigma(\cdot)$ is the activation function. Afterward, the updated node embeddings $\mathbf{S} = [\mathbf{s}_{1}, \mathbf{s}_{2}, \dots, \mathbf{s}_{|V|}]^{\top} \in \mathbb{R}^{|V|\times d}$ can be calculated as follows:
\begin{equation}
    \mathbf{S}=\mathbf{\Lambda} \cdot \mathbf{R}=\mathbf{\Lambda}\left(\sigma\left(\mathbf{U} \mathbf{\Lambda}^{\top} \mathbf{H}\right)+\mathbf{\Lambda}^{\top} \mathbf{H}\right).
\end{equation}

In this way, the updated node representations can effectively capture high-order semantic features. Finally, for each graph, the graph-level representation $\mathbf{s}_G$ can be obtained by summing all the refined node representations as:
\begin{equation}
    \mathbf{s}_G=\sum\nolimits_{i=1}^{|V|} \mathbf{s}_{i}.
\end{equation}

\subsection{Graph Convolution on the Line Graph}

Hypergraph modeling empowers our model to capture high-order semantic features and long-range dependencies. However, we argue that real-world graphs may involve interactions among higher-order substructures, whose interactions often reveal underlying meaningful semantic patterns. For instance, in biology, certain graph motifs may interact and collectively determine the properties of the graph~\cite{borgwardt2005protein,chen2006nemofinder}.

To this end, we leverage the learned hypergraph to introduce the line graph, effectively capturing interactions among hyperedges and providing a more profound exploration of the underlying semantic structure of the graph. Specifically, based on the hypergraph adjacency matrix $\mathbf{\Lambda} \in \mathbb{R}^{|V|\times k}$, we can construct the line graph $L(G)=\left(V_L, E_L\right)$ according to the Definition 3, where each node feature of the line graph is represented by the previously mentioned hyperedge embeddings $\mathbf{R} \in \mathbb{R}^{k \times d}$, and the adjacency matrix is denoted by $\mathbf{A}_L \in \mathbb{R}^{k \times k}$. Then, we can treat the line graph as a regular graph and adopt a GNN-based encoder to obtain its graph-level representation $\mathbf{t}_G$, which is calculated as:
\begin{equation}
    \begin{gathered}
        \mathbf{T}= \sigma(\hat{\mathbf{A}}_L \mathbf{R} \mathbf{W}), \quad \hat{\mathbf{A}}_L=\tilde{\mathbf{D}}_L^{-\frac{1}{2}} \tilde{\mathbf{A}}_L \tilde{\mathbf{D}}_L^{-\frac{1}{2}}, \\
        \mathbf{t}_G=\sum\nolimits_{i=1}^{k} \mathbf{t}_{i},
    \end{gathered}
\end{equation}
where $\mathbf{T}  = [\mathbf{t}_{1}, \mathbf{t}_{2}, \dots, \mathbf{t}_{k}]^{\top} \in \mathbb{R}^{k\times d}$, $\tilde{\mathbf{A}}_L=\mathbf{A}_L+\mathbf{I}$, $\tilde{\mathbf{D}}_L$ is the degree matrix of $\tilde{\mathbf{A}}_L$, $\mathbf{W}$ is the trainable weight matrix. In this way, the representation $\mathbf{t}_G$ can be viewed as another perspective of the original graph, considering interactions among high-order substructures and better capturing the underlying semantic structure.

\subsection{Relational Consistency Learning}

Having acquired the graph representations from the two perspectives, i.e., $\mathbf{s}_G$ from the hypergraph view and $\mathbf{t}_G$ from the line graph view, they capture semantic knowledge of the graph from different levels. More specifically, when we regard nodes within the same hyperedge as a substructure, the hypergraph convolution and line graph convolution channels within our network can be viewed as distinct perspectives describing intra-substructure and inter-substructure information. Thus, it is natural to consider how to integrate these two representations, enabling them to mutually supervise and reinforce each other. 

Furthermore, in semi-supervised scenarios, the availability of limited label annotations often leads to unreliable and biased pseudo labels. As a result, striving to align the graph representations or pseudo labels of the same instance directly might not prove to be the most effective strategy, especially for unlabeled graphs. To address this challenge, we suggest enhancing the representation of each instance by transfering knowledge among instances. This is achieved by comparing the similarities of the instance to other labeled graphs in the embedding spaces of the two branches.

Technically, we begin by randomly selecting a subset of labeled graphs $\{G_1,\dots,G_M\} \in \mathcal{G}^L$ as anchor graphs, which are stored in a memory bank. Then, we employ two branches to embed these anchor graphs, yielding the respective representations $\{\mathbf{s}_m\}_{m=1}^M$ and $\{\mathbf{t}_m\}_{m=1}^M$. To ensure that the anchor graphs sufficiently cover the neighborhoods of any unlabeled graph in the embedding space and facilitate the transfer of knowledge from labeled to unlabeled graphs, a large number of anchor graphs is required. However, processing excessive anchor graphs in a single iteration can become computationally expensive due to limitations in computation and memory resources. To address this challenge, we maintain a memory bank as a queue, which dynamically stores a set of anchor graphs selected from the most recent iterations of the two branches.
 
Specifically, take the hypergraph branch as an example, for an unlabeled graph $G_u$, we can calculate the relational similarity distribution between its graph representation $s_u$ with the representations $\left\{s_{m}\right\}_{m=1}^M$ of anchor graphs as:
\begin{equation}
    \mathcal{P}_u^m=\frac{\exp \left(\cos \left(\mathbf{s}_u, \mathbf{s}_{m}\right) / \tau\right)}{\sum_{m^{\prime}=1}^M \exp \left(\cos \left(\mathbf{s}_u, \mathbf{s}_{m^{\prime}}\right) / \tau\right)}.
\end{equation}
In accordance with \citet{you2020graph}, $\tau$ is the temperature parameter set to 0.5, $\cos(a,b)$ denotes the cosine similarity defined as $\frac{a \cdot b}{\left\|a\right\|_{2}\left\|b\right\|_{2}}$. Analogously, the relational similarity distribution in the line graph branch can be obtained as:
\begin{equation}
    \mathcal{Q}_u^m=\frac{\exp \left(\cos \left(\mathbf{t}_u, \mathbf{t}_{m}\right) / \tau\right)}{\sum_{m^{\prime}=1}^M \exp \left(\cos \left(\mathbf{t}_u, \mathbf{t}_{m^{\prime}}\right) / \tau\right)}.
\end{equation}

In this way, we propose the relational consistency learning to encourage the consistency between distributions $\mathcal{P}^u=\left[\mathcal{P}_1^u, \ldots, \mathcal{P}_M^u\right]$ and $\mathcal{Q}^u=\left[\mathcal{Q}_1^u, \ldots, \mathcal{Q}_M^u\right]$ by minimizing the Kullback-Leibler (KL) Divergence as:
\begin{equation}
\label{eq:con_loss}
    \mathcal{L}_{con}=\frac{1}{\left|\mathcal{G}^{U}\right|} \sum_{u \in \mathcal{G}^U} \frac{1}{2}\left(D_{\mathrm{KL}}\left(\mathcal{P}^u \| \mathcal{Q}^u\right)+D_{\mathrm{KL}}\left(\mathcal{Q}^u \| \mathcal{P}^u\right)\right).
\end{equation}

\noindent\textbf{Optimization Framework.} 
To introduce the supervision signals to guide the model, for each labeled graph $G_l$, we concatenate the graph representations $\mathbf{s}_l$ and $\mathbf{t}_l$ from the two branches and feed the fused representation to a classifier (multi-layer perception) for label prediction $\hat{\mathbf{y}}$. We then adopt cross-entropy loss to compute the supervised loss:
\begin{equation}
\label{eq:sup_loss}
    \mathcal{L}_{sup}=-\frac{1}{|\mathcal{G}^L|}\sum_{i \in \mathcal{G}^L }\mathbf{y}_i \log \left(\hat{\mathbf{y}}_i\right),
\end{equation}
where $\mathbf{y}_i$ denote the ground-truth label for labeled graph $G_i$. Finally, we integrate the supervised loss $\mathcal{L}_{sup}$ with relational consistency loss $\mathcal{L}_{con}$ in our combined loss:
\begin{equation}
\label{eq:total_loss}
\mathcal{L}=\mathcal{L}_{sup}+\beta \cdot \mathcal{L}_{con},
\end{equation}
where $\beta$ is a balance hyper-parameter. Our training algorithm is detailed in Algorithm~\ref{alg}.

\begin{algorithm}[t]
    \caption{Optimization Framework of the \method{}}
    \label{alg}
    \textbf{Input}: Labeled graphs $\mathcal{G}^L$, unlabeled graphs $\mathcal{G}^U$, the dimension of hidden embeddings $d$, the number of hyperedges $k$, the number of anchor graphs $M$, the hyper-parameter $\beta$ \\
    \textbf{Output}: Trained classifier
    \begin{algorithmic}[1]
        \STATE Initialize the parameters of the GNN-based encoder, hypergraph structure Learning, and classifier.\\ 
        \STATE Select $M$ anchor graphs from labeled set $\mathcal{G}^L$ to construct the memory bank.
        \WHILE{\emph{not convergence}}
            \STATE Sample a minibatch $\mathcal{B}^L$ and $\mathcal{B}^U$.
            \STATE Forward propagation $\mathcal{B}^L$ and $\mathcal{B}^U$ via twin branches.
            \STATE Compute consistency loss $\mathcal{L}_{con}$ by Eq.~\eqref{eq:con_loss}.
            \STATE Compute supervised loss $\mathcal{L}_{sup}$ by Eq.~\eqref{eq:sup_loss}.
            \STATE Update the parameters by gradient descent to minimize $\mathcal{L}$ by Eq.~\eqref{eq:total_loss}.
            \STATE Update the memory bank of the two branches following the first-in-first-out principle.
        \ENDWHILE 
    \end{algorithmic}
\end{algorithm}

\subsection{Computational Complexity Analysis}

With $\left| V \right|$ as the average number of nodes in input graphs, $d$ is the hidden dimensions, and $k$ is the hyperedge number, the total time complexity of obtaining hyperedge structure and performing hyperedge convolution is $O(k d (\left| V \right| + k ))$. And the time complexity of line graph convolution is $O(k d (k + d ))$. Moreover, the time complexity of computing relational consistency loss for a graph is $O(M d)$, where $M$ is the number of anchor graphs. Therefore, the total computational complexity of \method{} is $O(M d + k d (\left| V \right| + k + d) )$.


\section{Experiment}
\label{sec::experiment}

\begin{table*}[t]
\caption{Overview of performance (in \%) across six benchmark graph classification datasets, with standard deviations calculated over five runs. The highest scores are marked in bold, and the second-highest scores are underlined.}
\label{tab::results}
\centering
\tabcolsep=6.5pt
\begin{tabular}{lcccccc}
\toprule
{Methods} &  {PROTEINS} & {DD}   & {IMDB-B} & {IMDB-M}  & {REDDIT-M-5k} & {COLLAB} \\
\midrule 
WL  & $ 63.5\pm1.6 $ & $ 57.3\pm1.2 $  & $ 58.1\pm2.3  $ & $  33.3\pm1.4  $   & $ 37.0\pm0.9  $ & $ 62.9\pm0.7  $ \\
Sub2Vec & $ 52.7\pm4.5 $ & $ 46.4\pm3.2 $   & $ 44.9\pm3.5  $ & $  31.8\pm2.7  $  & $ 35.1\pm1.5  $ & $ 60.8\pm1.4  $ \\
Graph2Vec  & $ 63.1\pm1.8 $ & $ 53.7\pm1.6 $  & $ 61.2\pm2.6  $ & $  38.1\pm2.2  $   & $ 38.1\pm1.4  $ & $ 63.6\pm0.9  $ \\
\midrule
EntMin & $ 62.7\pm2.7 $ & $ 59.8\pm1.3 $    & $ 67.1\pm3.7 $ & $ 37.4\pm1.2 $  & $ 38.7\pm2.8 $ & $ 63.8\pm1.6 $  \\
Mean-Teacher  & $ 64.3\pm2.1 $ & $ 60.6\pm1.8 $   & $ 66.4\pm2.7 $ & $ 38.8\pm3.6 $  & $ 39.2\pm2.1 $  & $ 63.6\pm1.4 $\\
VAT  & $ 64.1\pm1.2 $ & $ 59.9\pm2.6 $   & $ 67.2\pm2.9 $ & $ 39.6\pm1.4 $  & $ 38.9\pm3.2 $ & $ 64.1\pm1.1 $ \\
\midrule
InfoGraph  & $ 68.2\pm0.7 $& $ 67.5\pm1.4 $  & $ 71.8\pm2.3 $ &  $ 42.3\pm1.8 $  & $ 41.5\pm1.7 $ & $ 65.7\pm0.4 $ \\
ASGN  & $ 67.7\pm1.2 $ & $ 68.5\pm0.6 $  & $ 70.6\pm1.4 $ & $ 41.2\pm1.4 $  & $ 42.2\pm0.8 $ & $ 65.3\pm0.8 $\\
GraphCL & $ 69.4\pm0.8 $ & $ 68.7\pm1.2 $ & $ 71.2\pm2.5 $ & $ 43.7\pm1.3 $ & $ 42.3\pm0.9 $ & $ 66.4\pm0.6 $\\
JOAO & $ 68.7\pm0.9 $ & $ 67.9\pm1.3 $ &  $ 71.0\pm1.9 $ & $ 42.6\pm1.5 $ & $ 42.1\pm1.2 $ & $ 65.8\pm0.4 $\\
DualGraph  &  $ 70.1\pm1.2 $  & $ 69.8\pm0.8 $  &  $ 72.1\pm0.7 $  &  \textbf{44.8 $\pm$ 0.4} &  $ 42.9\pm1.4 $  &  $ 67.2\pm0.6 $ \\
KGNN  &  $ 70.9\pm0.5 $  & $ 70.5\pm0.6 $  &  $ 72.5\pm1.6 $  &  43.3 $\pm$ 0.7 &  \underline{$ 44.8\pm0.6 $}  &  $ 67.4\pm0.5 $ \\
TGNN  &  \underline{$ 71.0\pm0.7 $}  &  \underline{$ 70.8\pm0.9 $}   &  \underline{$ 72.8\pm1.7 $}  &  42.9 $\pm$ 0.8  &  $ 43.8\pm1.0 $  &  \underline{$ 67.7\pm0.4 $} \\
\midrule 
\method{}  &  \textbf{73.4 $\pm$ 0.8}  &  \textbf{72.1 $\pm$ 0.9}   &  \textbf{73.5 $\pm$ 1.5}  &  \underline{44.3 $\pm$ 0.6}     &  \textbf{45.9 $\pm$ 1.0}  &  \textbf{68.3 $\pm$ 0.5} \\
\bottomrule
\end{tabular}
\end{table*}

\subsection{Experimental Setups}

\noindent\textbf{Datasets.}
We assess our \method{} on six publicly available datasets, comprising two bioinformatics datasets PROTEINS \cite{neumann2016propagation} and DD \cite{dobson2003distinguishing}; three datasets derived from social networks, specifically IMDB-B, IMDB-M, and REDDIT-M-5k \cite{yanardag2015deep}; and one dataset from scientific collaborations, COLLAB \cite{yanardag2015deep}. We employ the same data split with DualGraph ~\cite{luo2022dualgraph}, where the labeled training set, unlabeled training set, validation set, and test set are proportioned in a 2:5:1:2 ratio. Unless explicitly stated, we use 50\% of the labeled data (corresponding to 10\% of all samples) for training.

\noindent\textbf{Baseline Methods.} We conduct thorough comparisons with diverse methods categorized into three groups: traditional graph algorithms, conventional semi-supervised learning methods, and graph-specific semi-supervised learning approaches. Traditional graph methods include WL~\cite{shervashidze2011weisfeiler}, Graph2Vec~\cite{narayanan2017graph2vec}, and Sub2Vec~\cite{adhikari2018sub2vec}. Conventional semi-supervised learning approaches include EntMin~\cite{grandvalet2004semi}, Mean-Teacher~\cite{tarvainen2017mean} and VAT~\cite{miyato2018virtual}). The category of graph-specific semi-supervised learning methods encompasses InfoGraph~\cite{sun2020infograph}, GraphCL~\cite{you2020graph}, ASGN~\cite{hao2020asgn}, JOAO~\cite{you2021graph}), DualGraph~\cite{luo2022dualgraph}, KGNN~\cite{ju2022kgnn}, and TGNN~\cite{ju2023tgnn}.

\noindent\textbf{Implementation Details.}
For the implementation of \method{}, we employ the GIN \cite{xu2019powerful} to configure the GNN-based encoder. We empirically set the embedding dimension to 32, the batch size to 64, and the training epochs to 300. For our hypergraph structure learning module, we empirically set the number of hyperedge $k$ to 32. Moreover, we set the weight balance hyper-parameter $\beta$ for $\mathcal{L}_{con}$ to 0.01. The model \method{} is optimized using the Adam optimizer with an initial learning rate of 0.01, and the weight decay is set to 0.0005. Results are reported as the average classification accuracy (in $\%$) and the standard deviation over five runs.

\subsection{Results and Analysis}

The quantitative outcomes of semi-supervised graph classification are presented in Table \ref{tab::results}, and the following observations can be made from the results. (i) Traditional graph methods generally underperform compared to other methods, highlighting the superior capability of graph neural networks in harnessing valuable semantic information through advanced representation learning from graph-structured data. (ii) Graph-specific semi-supervised learning methods show enhanced performance over conventional semi-supervised learning techniques, demonstrating the suitability of recent graph semi-supervised learning for challenging graph classification tasks. Notably, KGNN and TGNN achieve nearly the best performance on most datasets, outperforming previous state-of-the-art approaches. The success of these approaches can be attributed to their proficient use of unlabeled samples, which boosts consistency across different modules in processing unlabeled graphs. (iii) Our proposed \method{} outperforms other baseline methods across the majority of benchmarks, demonstrating the robustness of our approach. The enhancement in performance can be attributed to the utilization of hypergraph and line graph convolution branches, which enable the capture of higher-order relationships among nodes. Additionally, the relational consistency learning module facilitates knowledge transfer between the two branches, leading to enhanced mutual guidance and improved performance. As for IMDB-M dataset, this discrepancy in performance can be attributed to the relatively smaller size of nodes (13.00) and edges (65.94) in the IMDB-M dataset compared to the others. In such cases, the use of hypergraph convolution may not be as effective, since the construction of a hypergraph might be unnecessary when dealing with a small-scale graph.

\begin{figure}[t]
\centering
\resizebox{1.0\linewidth}{!}{
    \subfigure[PROTEINS]{
     \includegraphics[width = 0.5\linewidth]{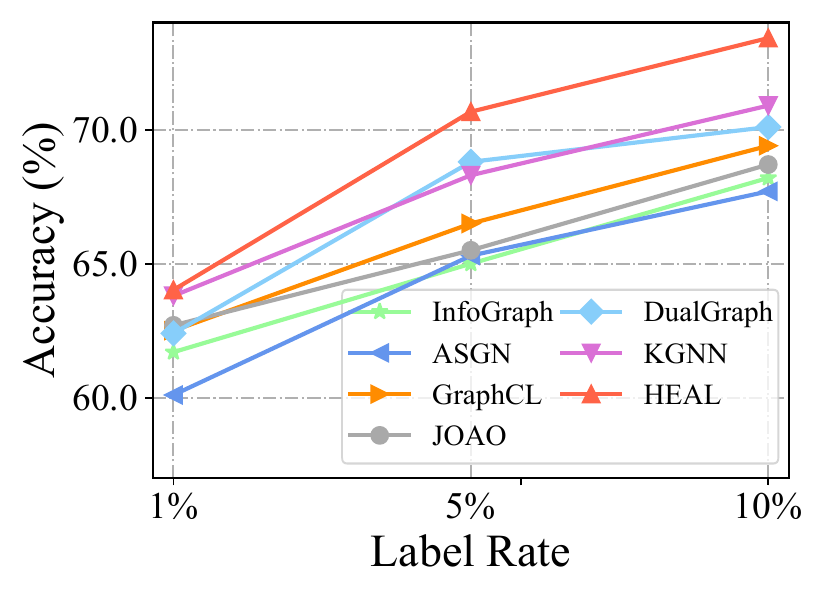}}
     \hfill
     \subfigure[REDDIT-M-5k]{
     \includegraphics[width = 0.5\linewidth]{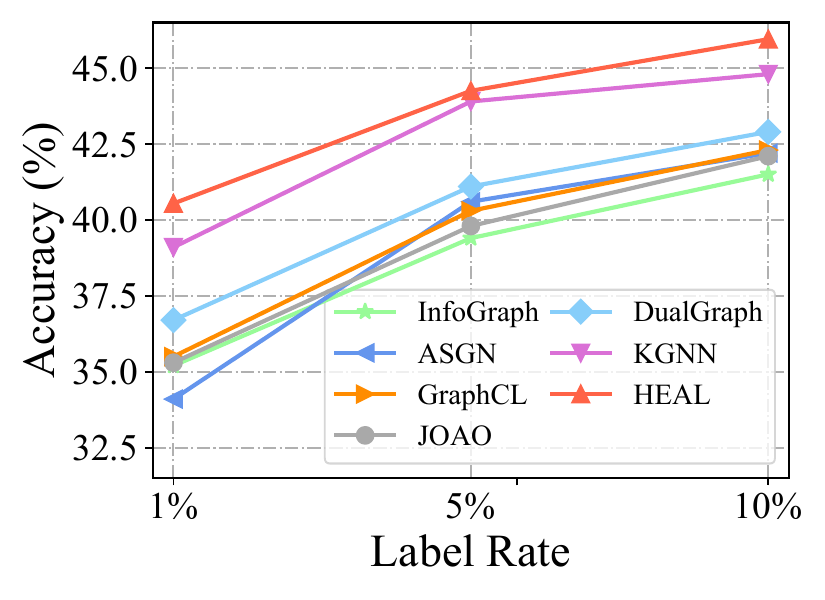}}
    }
\caption{Results of \method{} and baselines with different labeling ratios on PROTEINS and REDDIT-M-5k datasets.}
\label{fig:label_rate}
\end{figure}

\noindent\textbf{Influence of Labeling Ratio.} We evaluate our model \method{} and baselines on the PROTEINS and REDDIT-M-5k datasets by varying the labeling ratio of the training data, as shown in Figure \ref{fig:label_rate}. The findings illustrate that as the number of labeled instances rises, the performance of both \method{} and the baselines enhances, suggesting that adding more labeled data effectively boosts performance. Notably, our proposed \method{} demonstrates the best performance among all methods in most scenarios, underscoring the advantages of effectively incorporating graph semantics across different levels of higher-order structure.

\subsection{Ablation Study}

\begin{table}[t]
\caption{Ablation study of \method{} with several variants.}
\label{tab::ablation}
\renewcommand\arraystretch{0.9}
\centering
\tabcolsep=3.5pt
\resizebox{1.0\linewidth}{!}{
\begin{tabular}{lccc}
\toprule 
{Methods}   & {PROTEINS} & {REDDIT-M-5K} & {COLLAB} \\
\midrule 
Hyper-Sup  & $ 69.1\pm1.0 $  & $ 41.2\pm1.2 $ & $ 64.4\pm0.7 $ \\
Line-Sup  & $ 66.7\pm1.2 $  & $ 40.7\pm1.4 $  & $ 64.2\pm0.9 $ \\
Dual-Sup  & $ 70.1\pm0.9 $  & $ 42.4\pm1.2 $ & $ 65.3\pm0.8 $ \\
Hyper-Ensemble  & $ 71.8\pm1.1 $  & $ 43.6\pm1.3 $  & $ 66.7\pm0.7 $ \\
Line-Ensemble  & $ 71.5\pm1.1 $  & $ 43.9\pm1.4 $ & $ 66.0\pm0.8 $  \\
\midrule 
\specialrule{0em}{1pt}{1pt}
\method{}  &  \textbf{73.4 $\pm$ 0.8}   &  \textbf{45.9 $\pm$ 1.0}  &  \textbf{68.3 $\pm$ 0.5}  \\
\bottomrule
\end{tabular}
}
\end{table}

We carry out ablation studies to evaluate the impact of each component within our model. We test several model variants as outlined below:
(i) \textbf{Hyper-Sup} trains a single hypergraph convolution network using only supervised signals.
(ii) \textbf{Line-Sup} trains a single line graph convolution network in a supervised manner.
(iii) \textbf{Dual-Sup} trains a dual branch of the hypergraph and line graph convolution network in a supervised manner.
(iv) \textbf{Hyper-Ensemble} replaces the line graph convolution branch with another hypergraph learning module, using a different initialization.
(v) \textbf{Line-Ensemble} replaces hypergraph convolution branch with another line graph convolution module, also with different initialization.

The outcomes for various variants are displayed in Table~\ref{tab::ablation}. Firstly, it is evident that Hyper-Sup generally outperforms Line-Sup. 
Moreover, the combination of both hypergraph and line graph convolution (Dual-Sup) leads to improved performance, validating the joint effectiveness of both branches. Secondly, Hyper-Ensemble (Line-Ensemble) outperforms Hyper-Sup (Line-Sup), indicating that our relational consistency learning module effectively leverages unlabeled samples and improve the model via ensemble techniques. Lastly, our full model outperforms both ensemble versions, underscoring its enhanced effectiveness in extracting similarities by simultaneously considering both hypergraph and line graph perspectives.

\subsection{Hyper-parameter Study}

We analyze how the performance of \method{} changes with different hyper-parameter configurations. In particular, we assess the influence of the number of embedding dimensions $d$ and the number of hyperedges $k$ used within the hypergraph structure learning module.

\begin{figure}[t]
\centering
\resizebox{1.0\linewidth}{!}{
    \subfigure[Hyperedge Number]{
    \label{fig:hyperedge_num}
     \includegraphics[width = 0.5\linewidth]{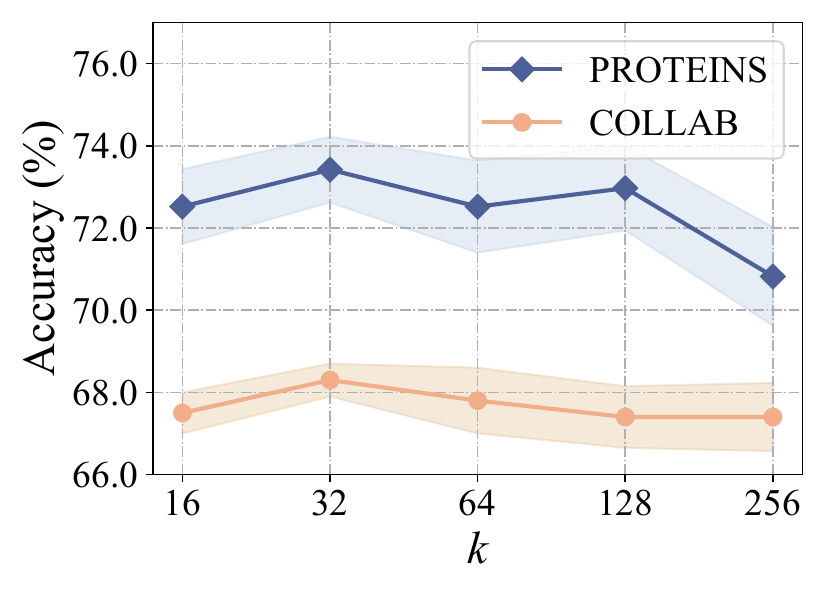}}
     \hfill
     \subfigure[Embedding Dimensions]{
     \label{fig:embedding_num}
     \includegraphics[width = 0.5\linewidth]{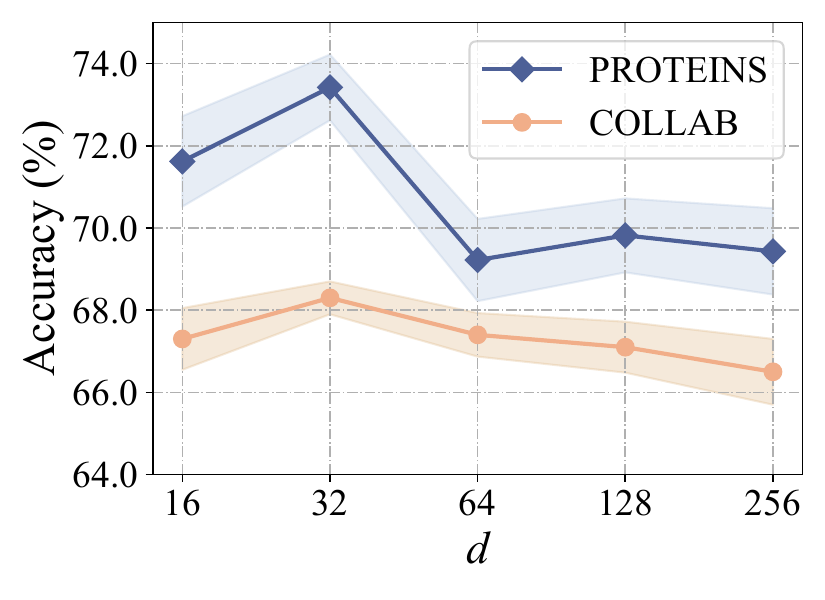}}
    }
\caption{Hyper-parameter sensitivity study of \method{} on PROTEINS and COLLAB datasets.}
\label{fig:hyperparam}
\end{figure}

\noindent\textbf{Influence of Hyperedge Numbers.} We first conducted the influence of the number of hyperedges, varying $k$ in $\{$16, 32, 64, 128, 256$\}$ while keeping all other hyperparameters fixed. The results are shown in Figure \ref{fig:hyperedge_num}, which reveal that the accuracy initially increases as the hyperedge number rises from 16 to 32. However, after reaching a peak, the accuracy starts to decrease with further increases in the number of hyperedges. This trend can be attributed to the potential capture of noise or aggregation of redundant information from hyperedges as the number of hyperedges grows.

\noindent\textbf{Influence of Embedding Dimensions.} We further evaluated the impact of embedding dimensions by varying $d$ in $\{$16, 32, 64, 128, 256$\}$ while maintaining all other hyper-parameters constant. The results depicted in Figure \ref{fig:embedding_num} indicate that performance reaches the peak when the embedding dimensions approach 32. This trend suggests that while increasing $d$ initially enhances the model's representation ability, it may result in overfitting if $d$ continues to rise.

\begin{figure}[t]
\centering
\resizebox{1.0\linewidth}{!}{
    \subfigure[PROTEINS]{
     \includegraphics[width = 0.5\linewidth]{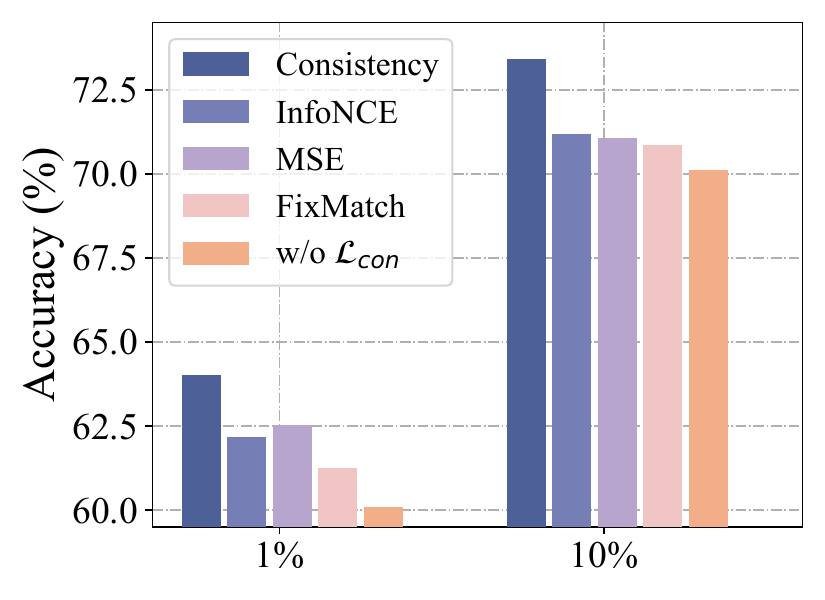}}
     \hfill
     \subfigure[COLLAB]{
     \includegraphics[width = 0.5\linewidth]{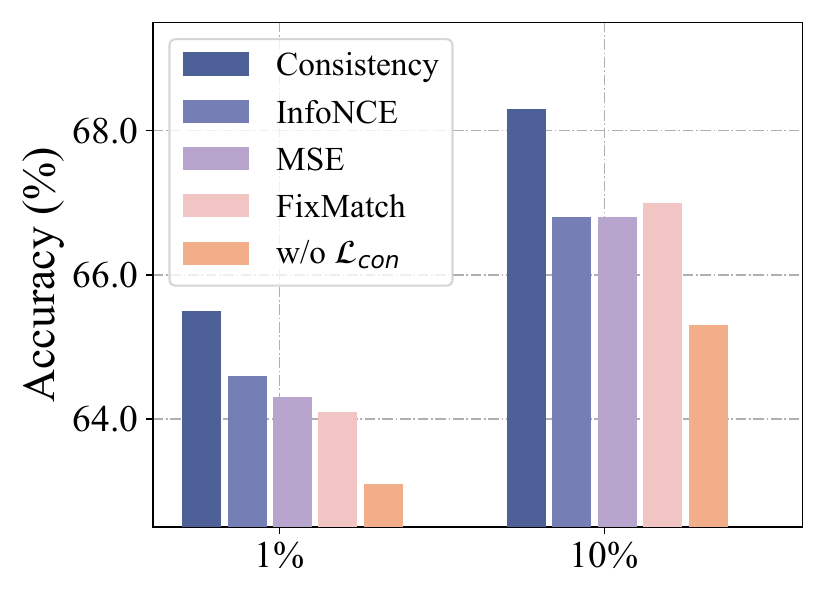}}
    }
\caption{Performance comparison with different labeling ratios w.r.t. different types of $\mathcal{L}_{con}$.}
\label{fig:loss}
\end{figure}

\subsection{Analysis of Consistency Loss}

The proposed relational consistency loss $\mathcal{L}_{con}$ aims to enhance each instance representation by exchanging instance knowledge from two correlated views. To highlight the advantages of the consistency learning module, we carry out experiments that compare $\mathcal{L}_{con}$ against other commonly employed contrastive losses (i.e. InfoNCE loss and mean squared error (MSE) loss) and consistency learning approaches (i.e., FixMatch loss). 
The results are presented in Figure \ref{fig:loss}, from which we can draw the following conclusions. Firstly, we observe a significant performance decline in \method{} when $\mathcal{L}_{con}$ is omitted (w/o $\mathcal{L}_{con}$), compared to its inclusion with various forms of contrastive loss. This observation highlights the importance of inter-branch knowledge communication in enhancing semi-supervised classification. Secondly, contrastive losses (Consistency, InfoNCE, MSE) generally surpass the pseudo-labeling consistency loss (FixMatch), which may be due to the biased pseudo-labels determined by unreliable prediction probabilities. 
Finally, our proposed consistency loss achieves better results than both InfoNCE and MSE. These methods typically concentrate on strictly enforcing similarities between two graph representations. Our findings suggest that a more flexible alignment of similarity distributions between hypergraph and line graph views enhances the effectiveness of consistency learning.

\subsection{Visualization Analysis}

\begin{figure}[t]
\centering
    \subfigure[Original Graph]{
    \label{fig:original}
     \includegraphics[width = 0.44\linewidth]{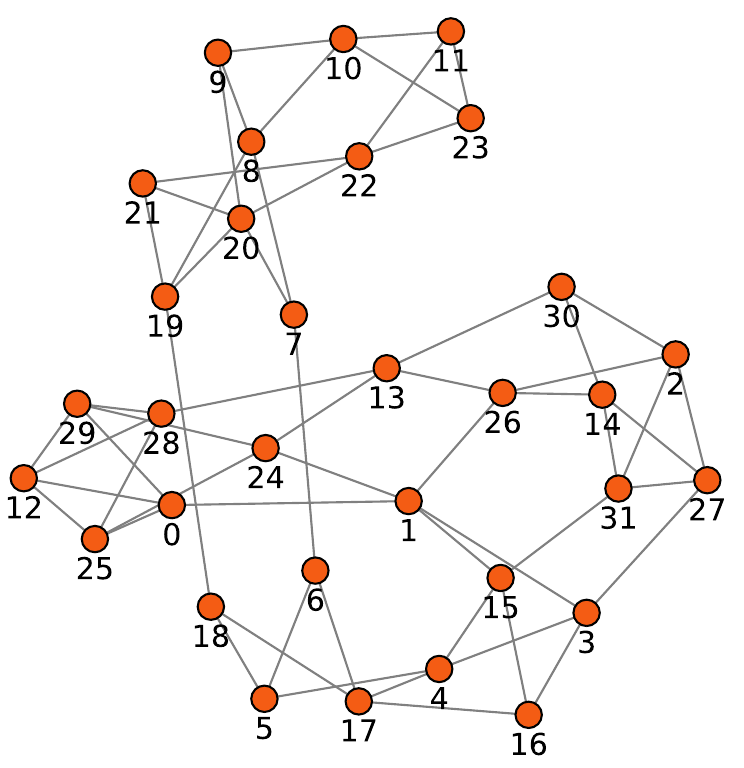}}
     \hfill
     \subfigure[Learned Hypergraph]{
     \label{fig:hypergraph}
     \includegraphics[width = 0.44\linewidth]{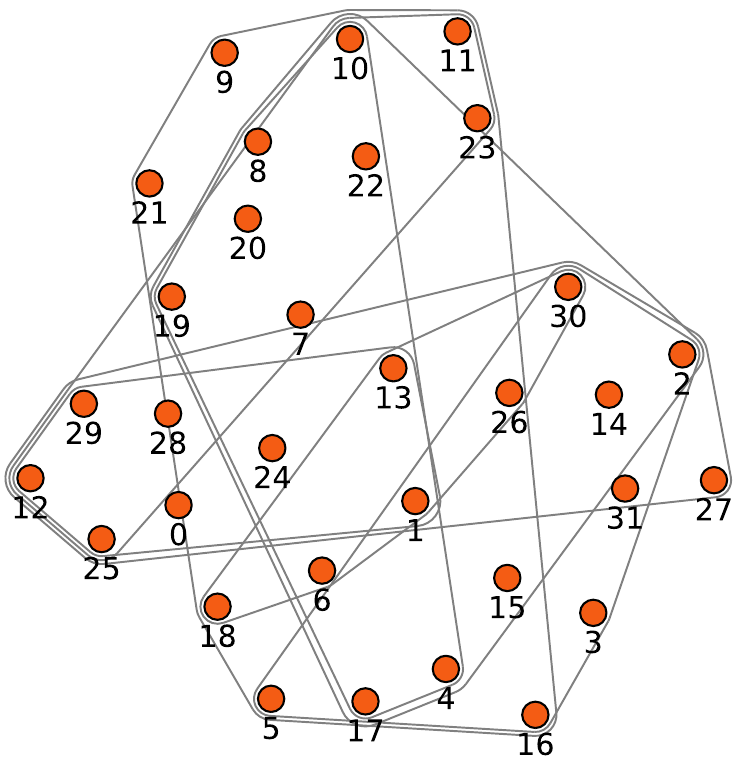}}\\
     
    \vspace{-2mm}
    \subfigure[Learned Line Graph]{
    \label{fig:line_graph}
     \includegraphics[width = 0.94\linewidth]{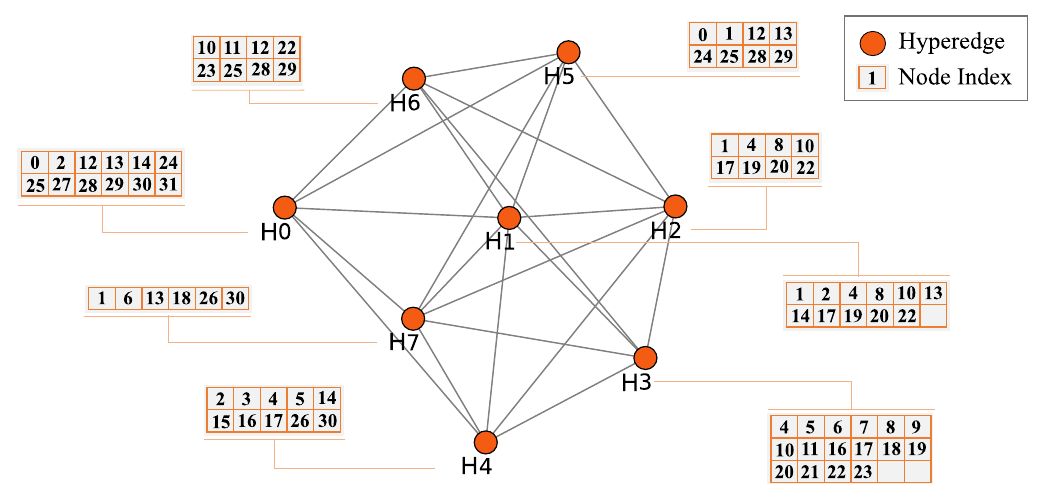}}
\caption{Visualization of the original graph, hypergraph structure and line graph learned by \method{} (only demonstrating the most significant hyperedges for simplicity).}
\label{fig:case}
\end{figure}

We conducted a case study on the PROTEINS dataset to visualize the learned hypergraph structure and corresponding line graph, thus demonstrating the effectiveness of the hypergraph structure learning module and line graph module, respectively. Here the thresholds for visualizing the learned hypergraph structure and corresponding line graph are both set to $0$.
In the PROTEINS dataset, each node represents secondary structure elements (helices, sheets, and turns), and the edges represent sequential or structural connections between nodes. Figure \ref{fig:original} reveals that elements in the protein are only connected to their nearest spatial neighbors, making it difficult to model higher-order interactions. However, in Figure \ref{fig:hypergraph}, we showcase part of the hyperedges learned by our hypergraph structure learning module. The hypergraph structure allows elements in the protein to interact in a high-order manner, facilitating the capturing of more complex and intricate relationships within protein structures. The results demonstrate that our hypergraph structure learning module exhibits remarkable adaptability in acquiring higher-order node relationships beyond pairwise connections, enabling enhanced flexibility in modeling complex data structures. Analogously, figure \ref{fig:line_graph} also depicts the effectiveness of the learned line graph, potentially involving interactions among hypergraphs, whose interactions often reveal underlying meaningful semantic patterns.

\section{Related Work}
\label{sec::related}


\noindent\textbf{Graph Neural Networks (GNNs)}
have risen as a powerful tool for handling graph-structured data, facilitating effective node and graph-level representation learning~\cite{ju2024survey}. The essence of GNNs lies in their iterative process of enhancing node representations by aggregating information from neighboring nodes, allowing nodes to propagate and exchange information throughout the graph~\cite{gilmer2017neural}. This message passing enables GNNs to capture local neighborhood information and learn expressive node representations, which are beneficial for a wide range of tasks such as node classification~\cite{yuan2023alex,luo2023towards_node}, node clustering~\cite{yi2023redundancy}, link prediction~\cite{zhang2018link,qin2024learning}, and graph classification~\cite{ju2022kgnn,luo2023rignn}. Compared with existing methods for supervised graph classification, our work goes further and studies a promising yet challenging semi-supervised graph classification.

\noindent\textbf{Hypergraph Learning}
have gained increasing attention for their ability to model complex relationships beyond pairwise interactions in traditional graphs. The underlying idea behind hypergraphs is to extend the concept of edges in graphs to hyperedges, which can connect multiple nodes simultaneously. This flexibility allows hypergraphs to capture higher-order dependencies and interactions among nodes. Various techniques have been proposed to leverage hypergraphs for diverse applications, including clustering~\cite{takai2020hypergraph}, classification~\cite{sun2021heterogeneous}, link prediction~\cite{yadati2020nhp}, traffic flow prediction~\cite{zhao2023dynamic}, knowledge graphs~\cite{fatemi2019knowledge}, and recommender systems~\cite{xia2021self}. Recently, hypergraph convolutional networks have been proposed as a generalization of GCNs to handle hypergraph-structured data, enabling effective feature aggregation and representation learning in hypergraphs~\cite{feng2019hypergraph,jiang2019dynamic,zhang2022deep,cai2022hypergraph}. Our \method{} also inherits the advantages of hypergraphs in modeling higher-order node relationships and additionally introduces a line graph to capture the semantic interactions between hyperedges.

\noindent\textbf{Semi-supervised Learning}
has been proven to be a prominent approach to address the limitations of traditional supervised learning, especially when labeled data is scarce or expensive to obtain. Early works in semi-supervised learning focus on spreading label knowledge from labeled data to neighboring unlabeled data points, effectively expanding the labeled set and providing more informative data points for training~\cite{subramanya2022graph,wan2021contrastive}. Another class involves consistency regularization~\cite{laine2017temporal,tarvainen2017mean,lucas2022barely}, which encourages the model to maintain stability in its predictions for perturbed versions of the same input, whether labeled or unlabeled. Compared with existing methods, our approach leverages the idea of hypergraphs for both labeled or unlabeled graphs to explore the inherent structure and relationships within the data.

\section{Conclusion}
\label{sec::conclusion}

In this work, we present a hypergraph-enhanced dual framework \method{} for semi-supervised graph classification, and our \method{} effectively captures graph semantics from the perspectives of hypergraph and line graph. It incorporates hypergraph structure learning to explore higher-order node dependencies and introduces a line graph to capture hyperedge interactions. Then, relational consistency learning is developed to facilitate knowledge transfer between the two branches. Experiments reveal superior performance compared to baseline methods in real-world graph datasets.

\section*{Acknowledge}
\vspace{-1mm}
This paper is partially supported by the National Natural Science Foundation of China (NSFC Grant Numbers 62306014 and 62276002) as well as the China Postdoctoral Science Foundation with Grant No. 2023M730057.

\section*{Impact Statement}
The proposed HEAL framework advances semi-supervised graph classification by incorporating hypergraph and line graph perspectives, addressing the limitations of traditional pairwise node relationships. By learning higher-order node dependencies through hypergraph structure learning and capturing hyperedge interactions via a line graph, HEAL enhances the extraction of underlying semantic structures. This dual approach facilitates improved knowledge transfer and mutual guidance between the two graph representations, contributing to more accurate and insightful graph classification. This work holds potential for broad applications in domains requiring effective and efficient graph analysis, such as social network analysis, biological networks, knowledge graphs, and recommender systems.

\bibliography{example_paper}

\begin{thebibliography}{55}
\providecommand{\natexlab}[1]{#1}
\providecommand{\url}[1]{\texttt{#1}}
\expandafter\ifx\csname urlstyle\endcsname\relax
  \providecommand{\doi}[1]{doi: #1}\else
  \providecommand{\doi}{doi: \begingroup \urlstyle{rm}\Url}\fi

\bibitem[Adhikari et~al.(2018)Adhikari, Zhang, Ramakrishnan, and Prakash]{adhikari2018sub2vec}
Adhikari, B., Zhang, Y., Ramakrishnan, N., and Prakash, B.~A.
\newblock Sub2vec: Feature learning for subgraphs.
\newblock In \emph{Advances in Knowledge Discovery and Data Mining: 22nd Pacific-Asia Conference, PAKDD 2018, Melbourne, VIC, Australia, June 3-6, 2018, Proceedings, Part II 22}, pp.\  170--182. Springer, 2018.

\bibitem[Borgwardt et~al.(2005)Borgwardt, Ong, Sch{\"o}nauer, Vishwanathan, Smola, and Kriegel]{borgwardt2005protein}
Borgwardt, K.~M., Ong, C.~S., Sch{\"o}nauer, S., Vishwanathan, S., Smola, A.~J., and Kriegel, H.-P.
\newblock Protein function prediction via graph kernels.
\newblock \emph{Bioinformatics}, 21\penalty0 (suppl\_1):\penalty0 i47--i56, 2005.

\bibitem[Cai et~al.(2022)Cai, Song, Sun, Zhang, Hong, and Li]{cai2022hypergraph}
Cai, D., Song, M., Sun, C., Zhang, B., Hong, S., and Li, H.
\newblock Hypergraph structure learning for hypergraph neural networks.
\newblock In \emph{Proceedings of the Thirty-First International Joint Conference on Artificial Intelligence}, pp.\  1923--1929, 2022.

\bibitem[Chen et~al.(2006)Chen, Hsu, Lee, and Ng]{chen2006nemofinder}
Chen, J., Hsu, W., Lee, M.~L., and Ng, S.-K.
\newblock Nemofinder: Dissecting genome-wide protein-protein interactions with meso-scale network motifs.
\newblock In \emph{Proceedings of the 12th ACM SIGKDD International Conference on Knowledge Discovery and Data Mining}, pp.\  106--115, 2006.

\bibitem[Dobson \& Doig(2003)Dobson and Doig]{dobson2003distinguishing}
Dobson, P.~D. and Doig, A.~J.
\newblock Distinguishing enzyme structures from non-enzymes without alignments.
\newblock \emph{Journal of Molecular Biology}, 330\penalty0 (4):\penalty0 771--783, 2003.

\bibitem[Fatemi et~al.(2019)Fatemi, Taslakian, Vazquez, and Poole]{fatemi2019knowledge}
Fatemi, B., Taslakian, P., Vazquez, D., and Poole, D.
\newblock Knowledge hypergraphs: Prediction beyond binary relations.
\newblock \emph{arXiv preprint arXiv:1906.00137}, 2019.

\bibitem[Feng et~al.(2019)Feng, You, Zhang, Ji, and Gao]{feng2019hypergraph}
Feng, Y., You, H., Zhang, Z., Ji, R., and Gao, Y.
\newblock Hypergraph neural networks.
\newblock In \emph{Proceedings of the AAAI Conference on Artificial Intelligence}, volume~33, pp.\  3558--3565, 2019.

\bibitem[Gilmer et~al.(2017)Gilmer, Schoenholz, Riley, Vinyals, and Dahl]{gilmer2017neural}
Gilmer, J., Schoenholz, S.~S., Riley, P.~F., Vinyals, O., and Dahl, G.~E.
\newblock Neural message passing for quantum chemistry.
\newblock In \emph{International Conference on Machine Learning}, pp.\  1263--1272. PMLR, 2017.

\bibitem[Grandvalet \& Bengio(2004)Grandvalet and Bengio]{grandvalet2004semi}
Grandvalet, Y. and Bengio, Y.
\newblock Semi-supervised learning by entropy minimization.
\newblock \emph{Advances in Neural Information Processing Systems}, 17:\penalty0 529--536, 2004.

\bibitem[Hao et~al.(2020)Hao, Lu, Huang, Wang, Hu, Liu, Chen, and Lee]{hao2020asgn}
Hao, Z., Lu, C., Huang, Z., Wang, H., Hu, Z., Liu, Q., Chen, E., and Lee, C.
\newblock Asgn: An active semi-supervised graph neural network for molecular property prediction.
\newblock In \emph{Proceedings of the 26th ACM SIGKDD International Conference on Knowledge Discovery \& Data Mining}, pp.\  731--752, 2020.

\bibitem[Huang et~al.(2015)Huang, Elhoseiny, Elgammal, and Yang]{huang2015learning}
Huang, S., Elhoseiny, M., Elgammal, A., and Yang, D.
\newblock Learning hypergraph-regularized attribute predictors.
\newblock In \emph{Proceedings of the IEEE Conference on Computer Vision and Pattern Recognition}, pp.\  409--417, 2015.

\bibitem[Jiang et~al.(2019)Jiang, Wei, Feng, Cao, and Gao]{jiang2019dynamic}
Jiang, J., Wei, Y., Feng, Y., Cao, J., and Gao, Y.
\newblock Dynamic hypergraph neural networks.
\newblock In \emph{Proceedings of the Twenty-Eighth International Joint Conference on Artificial Intelligence}, pp.\  2635--2641, 2019.

\bibitem[Ju et~al.(2022)Ju, Yang, Qu, Song, Shen, and Zhang]{ju2022kgnn}
Ju, W., Yang, J., Qu, M., Song, W., Shen, J., and Zhang, M.
\newblock Kgnn: Harnessing kernel-based networks for semi-supervised graph classification.
\newblock In \emph{Proceedings of the Fifteenth ACM International Conference on Web Search and Data Mining}, pp.\  421--429, 2022.

\bibitem[Ju et~al.(2023{\natexlab{a}})Ju, Liu, Qin, Feng, Wang, Guo, Luo, and Zhang]{ju2023few}
Ju, W., Liu, Z., Qin, Y., Feng, B., Wang, C., Guo, Z., Luo, X., and Zhang, M.
\newblock Few-shot molecular property prediction via hierarchically structured learning on relation graphs.
\newblock \emph{Neural Networks}, 163:\penalty0 122--131, 2023{\natexlab{a}}.

\bibitem[Ju et~al.(2023{\natexlab{b}})Ju, Luo, Qu, Wang, Chen, Deng, Hua, and Zhang]{ju2023tgnn}
Ju, W., Luo, X., Qu, M., Wang, Y., Chen, C., Deng, M., Hua, X.-S., and Zhang, M.
\newblock Tgnn: A joint semi-supervised framework for graph-level classification.
\newblock \emph{arXiv preprint arXiv:2304.11688}, 2023{\natexlab{b}}.

\bibitem[Ju et~al.(2024{\natexlab{a}})Ju, Fang, Gu, Liu, Long, Qiao, Qin, Shen, Sun, Xiao, et~al.]{ju2024comprehensive}
Ju, W., Fang, Z., Gu, Y., Liu, Z., Long, Q., Qiao, Z., Qin, Y., Shen, J., Sun, F., Xiao, Z., et~al.
\newblock A comprehensive survey on deep graph representation learning.
\newblock \emph{Neural Networks}, pp.\  106207, 2024{\natexlab{a}}.

\bibitem[Ju et~al.(2024{\natexlab{b}})Ju, Yi, Wang, Long, Luo, Xiao, and Zhang]{ju2024survey}
Ju, W., Yi, S., Wang, Y., Long, Q., Luo, J., Xiao, Z., and Zhang, M.
\newblock A survey of data-efficient graph learning.
\newblock \emph{arXiv preprint arXiv:2402.00447}, 2024{\natexlab{b}}.

\bibitem[Kashima et~al.(2003)Kashima, Tsuda, and Inokuchi]{kashima2003marginalized}
Kashima, H., Tsuda, K., and Inokuchi, A.
\newblock Marginalized kernels between labeled graphs.
\newblock In \emph{Proceedings of International Conference on Machine Learning}, pp.\  321--328, 2003.

\bibitem[Kipf \& Welling(2016)Kipf and Welling]{kipf2017semi}
Kipf, T.~N. and Welling, M.
\newblock Semi-supervised classification with graph convolutional networks.
\newblock \emph{arXiv preprint arXiv:1609.02907}, 2016.

\bibitem[Kojima et~al.(2020)Kojima, Ishida, Ohta, Iwata, Honma, and Okuno]{kojima2020kgcn}
Kojima, R., Ishida, S., Ohta, M., Iwata, H., Honma, T., and Okuno, Y.
\newblock kgcn: a graph-based deep learning framework for chemical structures.
\newblock \emph{Journal of Cheminformatics}, 12:\penalty0 1--10, 2020.

\bibitem[Laine \& Aila(2017)Laine and Aila]{laine2017temporal}
Laine, S. and Aila, T.
\newblock Temporal ensembling for semi-supervised learning.
\newblock In \emph{Proceedings of International Conference on Learning Representations}, 2017.

\bibitem[Li et~al.(2019)Li, Rong, Cheng, Meng, Huang, and Huang]{li2019semi}
Li, J., Rong, Y., Cheng, H., Meng, H., Huang, W., and Huang, J.
\newblock Semi-supervised graph classification: A hierarchical graph perspective.
\newblock In \emph{The World Wide Web Conference}, pp.\  972--982, 2019.

\bibitem[Lucas et~al.(2022)Lucas, Weinzaepfel, and Rogez]{lucas2022barely}
Lucas, T., Weinzaepfel, P., and Rogez, G.
\newblock Barely-supervised learning: Semi-supervised learning with very few labeled images.
\newblock In \emph{Proceedings of the AAAI Conference on Artificial Intelligence}, volume~36, pp.\  1881--1889, 2022.

\bibitem[Luo et~al.(2022)Luo, Ju, Qu, Chen, Deng, Hua, and Zhang]{luo2022dualgraph}
Luo, X., Ju, W., Qu, M., Chen, C., Deng, M., Hua, X.-S., and Zhang, M.
\newblock Dualgraph: Improving semi-supervised graph classification via dual contrastive learning.
\newblock In \emph{2022 IEEE 38th International Conference on Data Engineering (ICDE)}, pp.\  699--712. IEEE, 2022.

\bibitem[Luo et~al.(2023{\natexlab{a}})Luo, Ju, Gu, Qin, Yi, Wu, Liu, and Zhang]{luo2023towards_node}
Luo, X., Ju, W., Gu, Y., Qin, Y., Yi, S., Wu, D., Liu, L., and Zhang, M.
\newblock Towards effective semi-supervised node classification with hybrid curriculum pseudo-labeling.
\newblock \emph{ACM Transactions on Multimedia Computing, Communications and Applications}, 20, 2023{\natexlab{a}}.

\bibitem[Luo et~al.(2023{\natexlab{b}})Luo, Zhao, Mao, Qin, Ju, Zhang, and Sun]{luo2023rignn}
Luo, X., Zhao, Y., Mao, Z., Qin, Y., Ju, W., Zhang, M., and Sun, Y.
\newblock Rignn: A rationale perspective for semi-supervised open-world graph classification.
\newblock \emph{Transactions on Machine Learning Research}, 2023{\natexlab{b}}.

\bibitem[Luo et~al.(2023{\natexlab{c}})Luo, Zhao, Qin, Ju, and Zhang]{luo2023towards}
Luo, X., Zhao, Y., Qin, Y., Ju, W., and Zhang, M.
\newblock Towards semi-supervised universal graph classification.
\newblock \emph{IEEE Transactions on Knowledge and Data Engineering}, 36:\penalty0 416--428, 2023{\natexlab{c}}.

\bibitem[Mao et~al.(2023)Mao, Ju, Qin, Luo, and Zhang]{mao2023rahnet}
Mao, Z., Ju, W., Qin, Y., Luo, X., and Zhang, M.
\newblock Rahnet: Retrieval augmented hybrid network for long-tailed graph classification.
\newblock In \emph{Proceedings of the 31st ACM International Conference on Multimedia}, pp.\  3817--3826, 2023.

\bibitem[Miyato et~al.(2018)Miyato, Maeda, Koyama, and Ishii]{miyato2018virtual}
Miyato, T., Maeda, S.-i., Koyama, M., and Ishii, S.
\newblock Virtual adversarial training: a regularization method for supervised and semi-supervised learning.
\newblock \emph{IEEE Transactions on Pattern Analysis and Machine Intelligence}, 41\penalty0 (8):\penalty0 1979--1993, 2018.

\bibitem[Narayanan et~al.(2017)Narayanan, Chandramohan, Venkatesan, Chen, Liu, and Jaiswal]{narayanan2017graph2vec}
Narayanan, A., Chandramohan, M., Venkatesan, R., Chen, L., Liu, Y., and Jaiswal, S.
\newblock graph2vec: Learning distributed representations of graphs.
\newblock \emph{arXiv preprint arXiv:1707.05005}, 2017.

\bibitem[Neumann et~al.(2016)Neumann, Garnett, Bauckhage, and Kersting]{neumann2016propagation}
Neumann, M., Garnett, R., Bauckhage, C., and Kersting, K.
\newblock Propagation kernels: efficient graph kernels from propagated information.
\newblock \emph{Machine Learning}, 102\penalty0 (2):\penalty0 209--245, 2016.

\bibitem[Qin et~al.(2024)Qin, Ju, Wu, Luo, and Zhang]{qin2024learning}
Qin, Y., Ju, W., Wu, H., Luo, X., and Zhang, M.
\newblock Learning graph ode for continuous-time sequential recommendation.
\newblock \emph{IEEE Transactions on Knowledge and Data Engineering}, 2024.

\bibitem[Shervashidze et~al.(2009)Shervashidze, Vishwanathan, Petri, Mehlhorn, and Borgwardt]{shervashidze2009efficient}
Shervashidze, N., Vishwanathan, S., Petri, T., Mehlhorn, K., and Borgwardt, K.
\newblock Efficient graphlet kernels for large graph comparison.
\newblock In \emph{Proceedings of International Conference on Artificial Intelligence and Statistics}, pp.\  488--495, 2009.

\bibitem[Shervashidze et~al.(2011)Shervashidze, Schweitzer, Van~Leeuwen, Mehlhorn, and Borgwardt]{shervashidze2011weisfeiler}
Shervashidze, N., Schweitzer, P., Van~Leeuwen, E.~J., Mehlhorn, K., and Borgwardt, K.~M.
\newblock Weisfeiler-lehman graph kernels.
\newblock \emph{Journal of Machine Learning Research}, 12\penalty0 (9):\penalty0 2539--2561, 2011.

\bibitem[Subramanya \& Talukdar(2022)Subramanya and Talukdar]{subramanya2022graph}
Subramanya, A. and Talukdar, P.~P.
\newblock \emph{Graph-based semi-supervised learning}.
\newblock Springer Nature, 2022.

\bibitem[Sun et~al.(2020)Sun, Hoffmann, Verma, and Tang]{sun2020infograph}
Sun, F.-Y., Hoffmann, J., Verma, V., and Tang, J.
\newblock Infograph: Unsupervised and semi-supervised graph-level representation learning via mutual information maximization.
\newblock In \emph{Proceedings of International Conference on Learning Representations}, 2020.

\bibitem[Sun et~al.(2021)Sun, Yin, Liu, Chen, Cao, Shao, and Viet~Hung]{sun2021heterogeneous}
Sun, X., Yin, H., Liu, B., Chen, H., Cao, J., Shao, Y., and Viet~Hung, N.~Q.
\newblock Heterogeneous hypergraph embedding for graph classification.
\newblock In \emph{Proceedings of the 14th ACM International Conference on Web Search and Data Mining}, pp.\  725--733, 2021.

\bibitem[Takai et~al.(2020)Takai, Miyauchi, Ikeda, and Yoshida]{takai2020hypergraph}
Takai, Y., Miyauchi, A., Ikeda, M., and Yoshida, Y.
\newblock Hypergraph clustering based on pagerank.
\newblock In \emph{Proceedings of the 26th ACM SIGKDD International Conference on Knowledge Discovery \& Data Mining}, pp.\  1970--1978, 2020.

\bibitem[Tarvainen \& Valpola(2017)Tarvainen and Valpola]{tarvainen2017mean}
Tarvainen, A. and Valpola, H.
\newblock Mean teachers are better role models: Weight-averaged consistency targets improve semi-supervised deep learning results.
\newblock In \emph{Advances in Neural Information Processing Systems}, pp.\  1195--1204, 2017.

\bibitem[Wan et~al.(2021)Wan, Pan, Yang, and Gong]{wan2021contrastive}
Wan, S., Pan, S., Yang, J., and Gong, C.
\newblock Contrastive and generative graph convolutional networks for graph-based semi-supervised learning.
\newblock In \emph{Proceedings of the AAAI Conference on Artificial Intelligence}, volume~35, pp.\  10049--10057, 2021.

\bibitem[Wang et~al.(2015)Wang, Liu, and Wu]{wang2015visual}
Wang, M., Liu, X., and Wu, X.
\newblock Visual classification by $\ell_1 $-hypergraph modeling.
\newblock \emph{IEEE Transactions on Knowledge and Data Engineering}, 27\penalty0 (9):\penalty0 2564--2574, 2015.

\bibitem[Whitney(1992)]{whitney1992congruent}
Whitney, H.
\newblock Congruent graphs and the connectivity of graphs.
\newblock \emph{Hassler Whitney Collected Papers}, pp.\  61--79, 1992.

\bibitem[Xia et~al.(2021)Xia, Yin, Yu, Wang, Cui, and Zhang]{xia2021self}
Xia, X., Yin, H., Yu, J., Wang, Q., Cui, L., and Zhang, X.
\newblock Self-supervised hypergraph convolutional networks for session-based recommendation.
\newblock In \emph{Proceedings of the AAAI Conference on Artificial Intelligence}, volume~35, pp.\  4503--4511, 2021.

\bibitem[Xu et~al.(2019)Xu, Hu, Leskovec, and Jegelka]{xu2019powerful}
Xu, K., Hu, W., Leskovec, J., and Jegelka, S.
\newblock How powerful are graph neural networks?
\newblock In \emph{Proceedings of International Conference on Learning Representations}, 2019.

\bibitem[Yadati et~al.(2020)Yadati, Nitin, Nimishakavi, Yadav, Louis, and Talukdar]{yadati2020nhp}
Yadati, N., Nitin, V., Nimishakavi, M., Yadav, P., Louis, A., and Talukdar, P.
\newblock Nhp: Neural hypergraph link prediction.
\newblock In \emph{Proceedings of the 29th ACM International Conference on Information \& Knowledge Management}, pp.\  1705--1714, 2020.

\bibitem[Yanardag \& Vishwanathan(2015)Yanardag and Vishwanathan]{yanardag2015deep}
Yanardag, P. and Vishwanathan, S.
\newblock Deep graph kernels.
\newblock In \emph{Proceedings of the 21th ACM SIGKDD International Conference on Knowledge Discovery and Data Mining}, pp.\  1365--1374, 2015.

\bibitem[Yi et~al.(2023{\natexlab{a}})Yi, Ju, Qin, Luo, Liu, Zhou, and Zhang]{yi2023redundancy}
Yi, S., Ju, W., Qin, Y., Luo, X., Liu, L., Zhou, Y., and Zhang, M.
\newblock Redundancy-free self-supervised relational learning for graph clustering.
\newblock \emph{IEEE Transactions on Neural Networks and Learning Systems}, 2023{\natexlab{a}}.

\bibitem[Yi et~al.(2023{\natexlab{b}})Yi, Mao, Ju, Zhou, Liu, Luo, and Zhang]{yi2023towards}
Yi, S.-Y., Mao, Z., Ju, W., Zhou, Y.-D., Liu, L., Luo, X., and Zhang, M.
\newblock Towards long-tailed recognition for graph classification via collaborative experts.
\newblock \emph{IEEE Transactions on Big Data}, pp.\  1683--1696, 2023{\natexlab{b}}.

\bibitem[You et~al.(2020)You, Chen, Sui, Chen, Wang, and Shen]{you2020graph}
You, Y., Chen, T., Sui, Y., Chen, T., Wang, Z., and Shen, Y.
\newblock Graph contrastive learning with augmentations.
\newblock \emph{Advances in Neural Information Processing Systems}, 33:\penalty0 5812--5823, 2020.

\bibitem[You et~al.(2021)You, Chen, Shen, and Wang]{you2021graph}
You, Y., Chen, T., Shen, Y., and Wang, Z.
\newblock Graph contrastive learning automated.
\newblock In \emph{International Conference on Machine Learning}, pp.\  12121--12132. PMLR, 2021.

\bibitem[Yu et~al.(2012)Yu, Tao, and Wang]{yu2012adaptive}
Yu, J., Tao, D., and Wang, M.
\newblock Adaptive hypergraph learning and its application in image classification.
\newblock \emph{IEEE Transactions on Image Processing}, 21\penalty0 (7):\penalty0 3262--3272, 2012.

\bibitem[Yuan et~al.(2023)Yuan, Luo, Qin, Mao, Ju, and Zhang]{yuan2023alex}
Yuan, J., Luo, X., Qin, Y., Mao, Z., Ju, W., and Zhang, M.
\newblock Alex: Towards effective graph transfer learning with noisy labels.
\newblock In \emph{Proceedings of the 31st ACM International Conference on Multimedia}, pp.\  3647--3656, 2023.

\bibitem[Zhang \& Chen(2018)Zhang and Chen]{zhang2018link}
Zhang, M. and Chen, Y.
\newblock Link prediction based on graph neural networks.
\newblock \emph{Advances in Neural Information Processing Systems}, 31:\penalty0 5171--5181, 2018.

\bibitem[Zhang et~al.(2022)Zhang, Feng, Ying, and Gao]{zhang2022deep}
Zhang, Z., Feng, Y., Ying, S., and Gao, Y.
\newblock Deep hypergraph structure learning.
\newblock \emph{arXiv preprint arXiv:2208.12547}, 2022.

\bibitem[Zhao et~al.(2023)Zhao, Luo, Ju, Chen, Hua, and Zhang]{zhao2023dynamic}
Zhao, Y., Luo, X., Ju, W., Chen, C., Hua, X.-S., and Zhang, M.
\newblock Dynamic hypergraph structure learning for traffic flow forecasting.
\newblock In \emph{2023 IEEE 39th International Conference on Data Engineering (ICDE)}, pp.\  2303--2316. IEEE, 2023.

\end{thebibliography}
\bibliographystyle{icml2024}


\end{document}